\documentclass[]{article}
\pdfoutput=1

\usepackage{./proceed2e}

\usepackage{xcolor}
\usepackage{graphicx}
\usepackage{amsmath}
\usepackage{amsfonts}
\usepackage{amssymb}
\usepackage{amsthm}
\usepackage{enumerate}			
\usepackage{natbib}
\usepackage{comment}
\usepackage{mdwlist}			
\usepackage[ruled]{algorithm}
\usepackage{algpseudocode}
\usepackage[normalem]{ulem}

\bibpunct{(}{)}{;}{a}{,}{,}		

\theoremstyle{definition}
\newtheorem{dfn}{Definition}
\newtheorem{thm}{Theorem}
\newtheorem{prp}{Proposition}	
\newtheorem{cor}[prp]{Corollary}
\newtheorem{lem}[prp]{Lemma}

\newcommand{\indep}{\perp\!\!\!\perp}
\newcommand{\nindep}{\perp\!\!\!\perp\!\!\!\!\!\!\!\diagup\;}

\newcommand{\tet}{\! \relbar \! \relbar \!}

\newcommand{\tea}{\! \relbar \!\! \rightarrow \!}
\newcommand{\tem}{\! \relbar \!\!\! \relbar \!\! \ast \,}

\newcommand{\aet}{\! \leftarrow \!\! \relbar \!}

\newcommand{\aea}{\! \leftarrow \!\!\! \rightarrow \!}
\newcommand{\aem}{\! \leftarrow \!\! \ast \,}

\newcommand{\mea}{\,\ast \!\!\rightarrow \!}

\newcommand{\skel}{\cS}

\newcommand{\bfL}{\mathbf{L}}

\newcommand{\bfQ}{\mathbf{Q}}

\newcommand{\bfS}{\mathbf{S}}

\newcommand{\bfV}{\mathbf{V}}
\newcommand{\bfW}{\mathbf{W}}
\newcommand{\bfX}{\mathbf{X}}
\newcommand{\bfY}{\mathbf{Y}}
\newcommand{\bfZ}{\mathbf{Z}}

\newcommand{\I}{\mathcal{I}}
\newcommand{\G}{\mathcal{G}}

\newcommand{\M}{\mathcal{M}}
\newcommand{\cP}{\mathcal{P}}

\newcommand{\cS}{\mathcal{S}}
\newcommand{\AAxy}{AA(\{X,Y\})}
\newcommand{\bfAA}[1]{AA({#1})}

\newcommand{\bfZz}{\mathbf{Z}_{\setminus Z}}

\newcommand{\ci}[3]{{#1}\!\indep\!{#2}\,|\,{#3}}
\newcommand{\cd}[3]{{#1}\!\nindep\!{#2}\,|\,{#3}}

\newcommand{\mci}[3]{{#1}\!\indep\!{#2}\,|\,[{#3}]}
\newcommand{\mcii}[4]{{#1}\!\indep\!{#2}\,|\,{#3} \cup [{#4}]}

\newcommand{\mcdd}[4]{{#1}\!\nindep\!{#2}\,|\,{#3} \cup [{#4}]}
\newcommand{\path}[1]{\langle{#1}\rangle}

\title{Supplement - Learning Sparse Causal Models is not NP-hard} 
\author{ {\bf Tom Claassen$^*$, Joris M.~Mooij$^\ddagger$, and Tom Heskes$^*$} \\  
$^*$ Institute for Computer and Information Science, Radboud University Nijmegen\\ 
$^\ddagger$ Informatics Institute, University of Amsterdam \\
The Netherlands } 
 
\begin{document} 
 
\maketitle 

\begin{abstract} 
This article contains detailed proofs and additional examples related to the UAI-2013 submission `Learning Sparse Causal Models is not NP-hard'. The supplement follows the numbering in the main submission. 
\end{abstract} 

\setcounter{prp}{7}
\setcounter{dfn}{1}

\section{Preliminaries}
For reference purposes a few basic graphical model concepts, terms and definitions. For details, see e.g., \cite{RichSpir02}.

\subsection{Graphical model terminology}\label{sec:terminology}

A \textit{mixed graph} $\G$ is a graphical model that can contain three types of edges between pairs of nodes: directed ($\leftarrow$, $\rightarrow$), bidirected ($\leftrightarrow$), and undirected ($\relbar$). In this paper we only consider graphs with at most one edge between each pair of nodes, and with no node with an edge to itself.
If there is an edge $X \rightarrow Y$ in $\G$ then $X$ is a \textit{parent} of its \textit{child} $Y$, if $X \leftrightarrow Y$ then $X$ and $Y$ are \textit{spouses} of each other, and if $X \relbar Y$ then they are called \textit{neighbours}.
A \textit{path} $\pi = \path{X_1,X_2,\dots,X_n}$ is an ordered sequence of distinct nodes where each successive pair $(X_i,X_{i+1})$ along $\pi$ is adjacent (connected by an edge) in $\G$. A \textit{directed path} is a path of the form $X_1 \rightarrow X_2 \rightarrow \ldots \rightarrow X_n$. A \textit{directed cycle} is a directed path from $X_1$ to $X_n$ in combination with a directed edge $X_n \rightarrow X_1$. A \textit{directed acyclic graph} (DAG) is a graph that contains only directed edges, but has no directed cycle.
The \textit{skeleton} $\skel$ of a graph $\G$ is the undirected graph corresponding to the structure of $\M$, so that for each edge in $\G$ there is a undirected edge in $\skel$.
A node $X$ is an \textit{ancestor} of $Y$ (and $Y$ a \textit{descendant} of $X$) if there is a directed path from $X$ to $Y$ in $\G$, or $X = Y$. 
A vertex $Z$ is a \textit{collider} on a path $\pi = \path{\ldots,X,Z,Y,\ldots}$ if there are arrowheads at $Z$ on both edges from $X$ and $Y$, i.e., if $X \mea Z \aem Y$ (where the symbol $\ast$ stands for either an arrowhead mark or a tail mark), otherwise it is a \textit{noncollider}.
A \textit{trek} is a path without colliders.

In a DAG $\G$, a path $\pi = \path{X,\dots,Y}$ is said to be \textit{unblocked} relative to a set of vertices $\bfZ$, if and only if:
\vspace{-0.7cm}
\begin{enumerate*}
\item[(1)] every noncollider on $\pi$ is not in $\bfZ$, and
\item[(2)] every collider along $\pi$ is an ancestor of $\bfZ$,
\end{enumerate*}
\vspace{-0.3cm}
otherwise the path is \textit{blocked}. We say that a path $\pi = \path{X,\dots,Y}$ is \textit{blocked by node} $Z \in \bfZ$ iff $\pi$ is blocked given $\bfZ$, but unblocked relative to $\bfZ_{\setminus Z}$.
If there exists an unblocked path between $X$ and $Y$ relative to $\bfZ$ in $\G$ then $X$ and $Y$ are said to be \textit{d-connected} given $\bfZ$; if there is no such path then $X$ and $Y$ are \textit{d-separated} by $\bfZ$.

A mixed graph $\M$ is an \textit{ancestral graph} (AG) iff an arrowhead at $X$ on an edge to $Y$ implies that there is no directed path from $X$ to $Y$ in $\M$, and there are no arrowheads at nodes with undirected edges. As a result, arrowhead marks can be read as `is not an ancestor of', and all DAGs are ancestral. In an ancestral graph $\M$ a node $X$ is said to be \textit{anterior} to a node $Y$ if there is a so-called \textit{anterior path} from $X$ to $Y$ in $\M$ of the form $X \relbar \ldots \relbar (Z) \rightarrow \ldots \rightarrow Y$, possibly with $Z = X$ (no undirected part) or with $Z = Y$ (no directed part), or if $X = Y$. Arrowhead marks in an ancestral graph can therefore also be read as `is not anterior to'.
When applied to an ancestral graph \textit{d}-separation is also known as \textit{m}-separation.\footnote{Sometimes \textit{m}-connected is defined using `anterior' instead of `ancestor' in condition (2) of `unblocked', but as colliders have no undirected edges these two are equivalent.}
An ancestral graph is \textit{maximal} (MAG) if for any two non-adjacent vertices there is a set that separates them. 
A path $\pi$ between two nodes $(X,Y)$ in an ancestral graph is \textit{inducing with respect to a set of nodes $\bfZ$} iff every collider on $\pi$ is ancestor of $X$ or $Y$, and every noncollider is in $\bfZ$. Inducing paths w.r.t\ $\bfZ = \varnothing$ are called \textit{primitive}. 

Throughout the rest of this article, $X$, $Y$ and $\bfZ$ represent disjoint (subsets of) nodes (vertices, variables) in a graph, with sets denoted in boldface.
The set $Adj(X)$ refers to the nodes adjacent to $X$ in an AG $\M$, $An(X)$ represents the ancestors of $X$ in $\M$, and $Ant(X)$ the nodes anterior to $X$ in $\M$. Similar for sets, i.e., $X \in Adj(\bfZ)$ implies $\exists Z \in \bfZ: X \in Adj(Z)$; idem for $An(\bfZ)$ and $Ant(\bfZ)$. 

Every (maximal) ancestral graph $\M$ over nodes $\bfV$ corresponds to some underlying causal DAG $\G$ over variables $\bfV \cup \bfL \cup \bfS$, where the (possibly empty) sets of unobserved latent variables $\bfL$ and selection nodes $\bfS$ in $\G$ have been marginalized and conditioned out, see \citep{RichSpir02}. 
We denote the ancestors of $\bfW \subseteq \bfV \cup \bfL \cup \bfS$ in $\G$ as $An_{\G}(\bfW)$, where the subscript $\G$ highlights that the ancestorship relation is with respect to the underlying DAG $\G$ instead of $\M$. Pairs of nodes in $\bfV$ that share a common ancestor in the subgraph of $\G$ over $\bfL$ are said to be \textit{confounded}. Nodes in $An_{\G}(\bfS)$ are said to be subject to \textit{selection bias}. Both confounding and selection bias can give rise to links between nodes in a MAG, where confounding is associated with bidirected edges and selection bias with undirected edges.

The following helpful properties for reading ancestral information from a (M)AG $\M$ corresponding to an underlying causal DAG $\G$ are shown in \citep{RichSpir02}:
\begin{enumerate*}
\item[(1)] $X \tem Y \in \M \;\; \Rightarrow \;\; X \in An_\G(Y \cup \bfS)$,
\item[(2)] $X \aem Y \in \M \;\; \Rightarrow \;\; X \notin An_\G(Y \cup \bfS)$,
\item[(3)] $X \aea Y \in \M \,\; \Rightarrow \;\; \{X,Y\} \subset De_\G(\bfL)$,
\item[(4)] $X \tet Y \in \M \,\;\; \Rightarrow \;\; \{X,Y\} \subset An_\G(\bfS)$.
\end{enumerate*}
Conversely, a node subject to selection bias in $\G$ has no arrowhead in $\M$, and a node not subject to selection bias is not part of an undirected edge in $\M$.
Finally, note that $X \in An_{\G}(\bfY \cup \bfS) \iff X \in Ant(\bfY)$ for all $X \in \M$, $\bfY \subseteq \M$. 

The following definition is a special case of the definition in section 4.2.1 of \citep{RichSpir02}. Given an ancestral graph $\M$ over nodes $\bfV \cup \bfL$, the \emph{marginal MAG} $\M'$ over nodes $\bfV$ has the following edges:
$X, Y \in \bfV$ are adjacent in $\M'$ if there does not exist a set $\bfZ \subseteq \bfV \setminus \{X,Y\}$ that $m$-separates $X,Y$ in $\M$, and in that case
the edge in $\M'$ has an arrowhead at $X$ if and only if $X \not\in Ant_{\M}(Y)$, and has an arrowhead at $Y$ if and only if $Y \not\in Ant_{\M}(X)$.

\subsection{Ancestral graph properties}
We rely on the following connection between in/dependences and (non-)ancestorship in an ancestral graph.

\textbf{Lemma 2.}
For disjoint (subsets of) nodes $X,Y,Z, \bfZ$ in an ancestral graph $\M$, 
\vspace{-0.3cm}
\begin{enumerate}
\item[(1)] $\mcdd{X}{Y}{\bfZ}{Z} \;\;\;\Rightarrow \;\; Z \notin Ant(\{X,Y\} \cup \bfZ)$.
\item[(2)] $\mcii{X}{Y}{\bfZ}{Z} \;\;\;\Rightarrow \;\; Z \in Ant(\{X,Y\} \cup \bfZ)$.
\item[(3)] $\mci{X}{Y}{\bfZ \cup Z} \;\;\;\Rightarrow \;\; Z \in Ant(\{X,Y\})$,
\end{enumerate}
\vspace{-0.3cm}
where square brackets indicate a \textit{minimal} set of nodes.
\vspace{-0.6cm}
\begin{proof}
See e.g., Corollary to Lemma 14 in \citep{SpirMR99}, and Lemma 2 in \citep{TomUAI11}. 
\end{proof}

We use the following result on anteriorship for nodes on unblocked paths:

\textbf{Lemma 3.13} 
In an ancestral graph $\M$, if $\pi$ is a path \textit{m}-connecting $X$ and $Y$ given $\bfZ$, then every vertex on $\pi$ is in $Ant(\{X,Y\} \cup \bfZ)$.
\begin{proof}
See 3.13 in \citep{RichSpir02}.
\end{proof}

As a result, rule (3) in Lemma 2 not only applies to the nodes in the minimal separating set, but also to \textit{all other nodes} on the paths in $\M$ between $X$ and $Y$ that become unblocked given only a subset $\bfZ^\prime \subset \bfZ$.
\begin{cor} \label{corMCIPathCaus}	
Let $\M$ be an ancestral graph, and suppose that $\mci{X}{Y}{\bfZ}$. If a path $\pi$ in $\M$ between $X$ and $Y$ is unblocked given some subset $\bfZ^\prime \subset \bfZ$,
then all nodes on $\pi$ are in $Ant(\{X,Y\})$.
\begin{proof}
Follows from Lemma 3.13, given that Lemma 2 rule (3) ensures that $\bfZ^\prime \subset \bfZ \subseteq Ant(\{X,Y\})$. 
\end{proof}
\end{cor}

Similarly, we can freely add anterior nodes to any separating set without introducing a dependence:
\begin{cor}  \label{corAAsep}	
In an ancestral graph $\M$, if $\ci{X}{Y}{\bfZ}$, then $\forall \bfW \subseteq Ant(\{X,Y\} \cup \bfZ)_{\setminus \{X,Y\}}: \ci{X}{Y}{\bfZ \cup \bfW}$.
\begin{proof}
Adding the nodes in $\bfW$ to the separating set one by one, then by rule (1) in Lemma 2, any node that creates a dependence cannot be anterior to any node in $\{X,Y\} \cup \bfZ$, contrary the assumed. So all added nodes leave the original independence intact, and therefore $\ci{X}{Y}{\bfZ \cup \bfW}$.
\end{proof}
\end{cor}

\section{\textit{D}-separating sets}
This part contains the proofs for section \S4.1 in the main article. We start by formalizing some terminology on \textit{D}-separation:

\begin{dfn} \label{dfnPDSep}
In a MAG $\M$, two nodes $X$ and $Y$ are \textbf{\textit{D}-separated} by a set of nodes $\bfZ$ iff:
\vspace{-0.2cm}
\begin{enumerate}
\item $\ci{X}{Y}{\bfZ}$, 
\item $\forall \bfZ^\prime \subseteq Adj(\{X,Y\})_{\setminus \{X,Y\}}: \cd{X}{Y}{\bfZ^\prime}$. 
\end{enumerate}
\vspace{-0.2cm}
If $\bfZ$ \textit{D}-separates $X$ and $Y$, then $(X,Y)$ is called a \textbf{\textit{D}-sep link}, and a node $Z \in \bfZ$ is called a \textbf{\textit{D}-sep node} for $(X,Y)$ if:
\vspace{-0.2cm}
\begin{enumerate}
\item $Z \notin Adj(\{X,Y\})$,
\item $\forall \bfZ^\prime \subseteq Adj(\{X,Y\}): \cd{X}{Y}{\bfZ_{\setminus Z} \cup \bfZ^\prime}$.
\end{enumerate}
\vspace{-0.2cm}
\end{dfn}
In words: $X$ and $Y$ are \textit{D}-separated by $\bfZ$ iff they are \textit{d}-separated by $\bfZ$, and all sets that \textit{can} separate $X$ and $Y$ contain at least one node $Z \notin Adj(\{X,Y\})$. 
Such a node $Z \in \bfZ$ that cannot be made redundant by nodes adjacent to $X$ or $Y$ is a \textit{D}-sep node, and the relation between $X$ and $Y$ is called a \textit{D}-sep link.

\subsection{Identifying \textit{D}-sep links}

To prove Lemma 3 from the main article we first derive a connection between `not separable by adjacent nodes' and non-anteriorship: 
\begin{lem}  \label{lemNonAdjSepNonCaus}
In an ancestral graph $\M$, if $\ci{X}{Y}{\bfZ}$, but $X$ is not independent of $Y$ given any subset of $Adj(X)$ in $\M$, then $Y \notin Ant(X)$ and $Y$ is not part of an undirected edge.
\begin{proof}
From $\ci{X}{Y}{\bfZ}$ there is no edge between $X$ and $Y$ in $\M$.
Let $\bfW = Adj(X) \cap Ant(\{X,Y\})$ be the set of all nodes in $\M$ that are adjacent to $X$ and have an anterior path to $X$ and/or $Y$. According to the assumed then $\cd{X}{Y}{\bfW}$, and so
there are one or more unblocked paths of the form $\path{X,U,\dots,Y}$ relative to $\bfW$ in $\M$ (as there is no direct edge). By Lemma 3.13 we know that implies $U \in Ant(\{X,Y\} \cup \bfW)$.
From $\bfW \subset Ant(\{X,Y\})$ and transitivity of `anteriorship' then follows $U \in Ant(\{X,Y\}$, which combined with the fact that $U$ is adjacent to $X$ implies $U \in \bfW$.

But given that path $\pi = \path{X,U,\dots,Y}$ is unblocked relative to $\bfW$, node $U \in \bfW$ must be a collider along this path with arrowhead $X \mea U$ in $\M$.
This means \mbox{$U \notin Ant(X)$}, which leaves $U \in Ant(Y)$. But then also $Y \notin Ant(X)$, otherwise (again by transitivity) $U$ would still be anterior to $X$ in $\M$. 
From the fact that $U$ is collider along $\pi$ we know that it is not part of an undirected edge, and so $Y$ as descendant of $U$ also cannot be part of an undirected edge in $\M$.
\end{proof}
\end{lem}
This also applies directly to \textit{D}-sep links.

\textbf{Lemma 3.}
In a MAG $\M$, if two nodes $X$ and $Y$ are \textit{D}-separated by a minimal set $\bfZ$, then
\vspace{-0.3cm}
\begin{enumerate*}
\item[(1)] $X \notin Ant(Y \cup \bfZ)$,
\item[(2)] $Y \notin Ant(X \cup \bfZ)$,
\item[(3)] $\forall Z \in \bfZ: Z \in Ant(\{X,Y\})$,
\item[(4)] $X$ and $Y$ are not part of an undirected edge.
\end{enumerate*}
\vspace{-0.3cm}
\begin{proof}
(1) from the definition of \textit{D}-separated nodes and Lemma \ref{lemNonAdjSepNonCaus} follows that $X$ is not anterior to $Y$; but $X$ also cannot be anterior to any node in $\bfZ$, otherwise by (3) and transitivity/acyclicity it would either still be anterior to $Y$, or it would by anterior to itself which would imply a directed cycle (given that (4) implies there cannot be an undirected edge at $X$); therefore $X \notin Ant(Y \cup \bfZ)$;

(2) idem for $Y$;

(3) Lemma 2 rule (3), given that $\bfZ$ is minimal.

(4) follows directly from Lemma \ref{lemNonAdjSepNonCaus}.
\end{proof}
Note that it is possible that one or more nodes in $\bfZ$ (including \textit{D}-sep nodes) are part of an undirected edge in $\M$.

Next we introduce:\\
\begin{dfn} \label{dfnAA}
For a set of nodes $\bfX$ in an ancestral graph $\M$, the set $\bfAA{\bfX}$ (\textit{adjacent anteriors}) is defined as $\bfAA{\bfX} = (Adj(\bfX) \cap Ant(\bfX)) \setminus \bfX$.
\end{dfn}
In the context of \textit{D}-sep links $(X,Y)$ we usually refer to $\AAxy$ as the set of \textit{adjacent ancestors}, as then $Ant(\{X,Y\}) = An(\{X,Y\})$, by Lemma 3-(4).

With this we can bring \textit{D}-separation in standard form:
\begin{lem}  \label{lemDsepNodes}	
In a MAG $\M$, if two nodes $X$ and $Y$ are \textit{d}-separated by $\bfZ$, then also $\mci{X}{Y}{\bfZ_{AA} \cup \bfZ_{DS}}$, with $\bfZ_{DS} \subset \bfZ$, $\bfZ_{AA} \subseteq \AAxy$, $\bfZ_{AA} \cap \bfZ_{DS} = \varnothing$, and where all nodes in $\bfZ_{DS}$ (possibly empty) are \textit{D}-sep nodes for $(X, Y)$.
\begin{proof}
We use rules (1)-(3) in Lemma 2 to construct the two sets. First we remove nodes from $\bfZ$ one-by-one until no more can be removed to obtain a minimal $\mci{X}{Y}{\bfZ'}$, with $\bfZ' \subseteq \bfZ$.
By rule (3), all nodes in $\bfZ'$ are anterior to $X$ and/or $Y$.
By Corollary \ref{corAAsep} we obtain \mbox{$\ci{X}{Y}{\AAxy \cup \bfZ''}$}, where $\bfZ'' = \bfZ' \setminus \AAxy$ contains the subset of nodes from $\bfZ'$ that are not adjacent to $X$ and/or $Y$.

We obtain $\bfZ_{DS}$ by eliminating nodes from $\bfZ''$ one by one until no more nodes can be eliminated without destroying the independence, and so then \mbox{$\mcii{X}{Y}{\AAxy}{\bfZ_{DS}}$}. 
If $(X,Y)$ is not a \textit{D}-sep link, then $\bfZ_{DS} = \varnothing$. 
Finally we can obtain $\bfZ_{AA}$ by eliminating superfluous nodes from $\AAxy$ one by one until no more can be removed without creating a dependence. 

By construction the sets $\bfZ_{AA}$ and $\bfZ_{DS}$ are disjoint.
No additional nodes from $\bfZ_{DS}$ can be eliminated during/after the process of eliminating nodes from $\AAxy$: if $Z \in \bfZ_{DS}$ can be eliminated only \textit{after} some node $Z_A \in \AAxy$ is eliminated, then putting back $Z_A$ after $Z$ is removed should create a dependence, in contradiction with Corollary \ref{corAAsep}. 
Therefore, at that point the \textit{D}-separating set is minimal, i.e., $\mci{X}{Y}{\bfZ_{AA} \cup \bfZ_{DS}}$. 

All nodes in $\bfZ_{DS}$ (if nonempty) satisfy the definition of \textit{D}-sep node: by construction none of them are adjacent to $X$ or $Y$, and if there were some subset $\bfW \subseteq Adj(\{X,Y\})$ that could make a node $Z \in \bfZ_{DS}$ redundant then, by Lemma 2.(2), that subset $\bfW$ must be a subset of $\AAxy$, and so by Corollary \ref{corAAsep} the independence should also be found given $\AAxy \cup \bfZ'''$ with $\bfZ''' \subseteq \bfZ_{DS} \setminus \{Z\}$: a contradiction.
\end{proof}
\end{lem}
Note that neither $\bfZ_{AA}$ nor $\bfZ_{DS}$ need be uniquely defined for a given \textit{D}-separated $\mci{X}{Y}{\bfZ}$, but may depend on the order in which nodes are removed.

\begin{figure}[h]
\begin{center}
\includegraphics[scale=0.9]{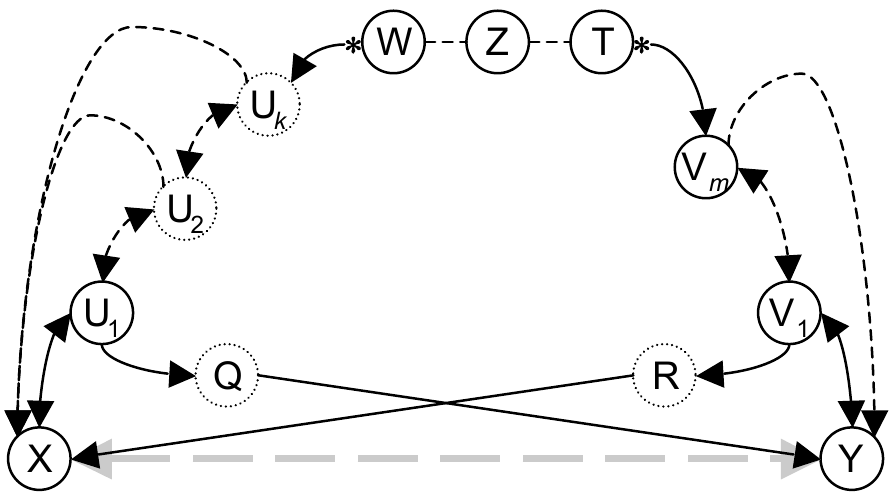}
\caption{\small{ Path configuration for \textit{D}-sep link $X - Y$.}} \label{figDSepEdgePath}
\end{center}
\end{figure}

In the proof of Lemma 4 we rely on the fact that for each \textit{D}-sep link there is a path blocked by a \textit{D}-sep node of the form depicted in Figure \ref{figDSepEdgePath}, which imposes six \textit{identifiable} minimal dependence relations in (4)-(6), below:

\begin{lem}  \label{lemPD_CausID}
  In a MAG $\M$, if nodes $X$ and $Y$ are \textit{D}-separable, then there are nodes $\{U,V,W,T\} \in An(\{X,Y\})$ such that:
\begin{enumerate}
\item[(1)] $X \aea U$ and $V \aea Y$ in $\M$,
\item[(2)] $U \in An(Y)$ and $V \in An(X)$,
\item[(3)] $U \notin An(V)$ and $V \notin An(U)$,
\item[(4)] $W \notin Adj(X)$ and $\forall \bfZ_{XW}$ with $\mci{X}{W}{\bfZ_{XW}}$:\\ $\mcdd{X}{W}{\bfZ_{XW}}{U}$ and $\mcdd{X}{W}{\bfZ_{XW}}{Y}$,
\item[(5)] $T \notin Adj(Y)$ and $\forall \bfZ_{YT}$ with $\mci{Y}{T}{\bfZ_{YT}}$:\\ $\mcdd{Y}{T}{\bfZ_{YT}}{V}$ and $\mcdd{Y}{T}{\bfZ_{YT}}{X}$,
\item[(6)] $U \notin Adj(V)$ in $\M$, and $\forall \bfZ_{UV}, \mci{U}{V}{\bfZ_{UV}}$:\\
 $\mcdd{U}{V}{\bfZ_{UV}}{X}$ and $\mcdd{U}{V}{\bfZ_{UV}}{Y}$.
\end{enumerate}
\begin{proof}
By Lemma \ref{lemDsepNodes} we have $\mci{X}{Y}{\bfZ_{AA} \cup \bfZ_{DS}}$, with $\bfZ_{AA} \subseteq \AAxy$, and $\bfZ_{DS}$ a (sub)set of \textit{D}-sep nodes not adjacent to $X$ and/or $Y$.
Let $Z \in \bfZ_{DS}$ and define $\bfZ := \AAxy \cup \bfZ_{DS}$. Then, $\ci{X}{Y}{\bfZ}$ but $\cd{X}{Y}{\bfZ_{\setminus Z}}$, and so there must be a path $\pi$ that is (only) blocked by noncollider $Z$ (relative to the other $\bfZ_{\setminus Z}$). 

We now show that we can take this path to be of the form $\pi = X \leftrightarrow U_1 (\leftrightarrow U_2 \leftrightarrow .. \leftrightarrow U_k) \aem W \cdots Z \cdots T \mea (V_m \leftrightarrow .. \leftrightarrow V_2 \leftrightarrow) V_1 \leftrightarrow Y$ 
in $\M$, where all nodes $U_i$ are colliders along $\pi$ and are adjacent to $X$, but only $U_1$ has a bidirected edge to $X$ (similar for $V_i$ at $Y$), and $W$ is the first node along $\pi$ starting from $X$ that is not adjacent to $X$ (possibly $W = Z$), and similarly for $T$. See also Figure \ref{figDSepEdgePath}. 

Firstly, all paths between $X$ and $Y$ blocked by a node $Z \in \bfZ_{DS}$ 
must be \textit{into} both $X$ and $Y$: given that there are no undirected edges to $X$ and/or $Y$ in $\M$ (Lemma 3), then by Corollary \ref{corMCIPathCaus} the first node $U$ encountered along any such path must be in $An(\{X,Y\})$. But if this path starts with a tail from $X$ then necessarily $X \tea U$, so that $U \in An(Y)$, which in turn implies $X \in An(Y)$, in contradiction with Lemma 3. Idem for $Y$.
Therefore all paths blocked by node $Z$, including $\pi$, must have $X \aem \ldots \mea Y$.

Secondly, all paths between $X$ and $Y$ blocked by a node $Z \in \bfZ_{DS}$ must go via at least two other nodes $U_1 \in Adj(X)$ resp.\ $V_1 \in Adj(Y)$, as $Z$ is presumed to be not adjacent to $X$ and/or $Y$. As both these nodes satisfy the criteria for $\AAxy$ they are part of the conditioning set $\bfZ$, and so they must be colliders along $\pi$ (otherwise $Z$ was not needed to block it). The same holds for all subsequent nodes up to $U_k$ and $V_m$ along $\pi$ that are adjacent to $X$ and/or $Y$.
Therefore the path $\pi$ blocked by $Z$ must have the general form $\pi = X \leftrightarrow U (\leftrightarrow \ldots \leftrightarrow U') \aem \ldots \mea (V' \leftrightarrow \ldots \leftrightarrow) V \leftrightarrow Y$.

Next, starting from $X$, at some point along $\pi$ the first node $W$ must be encountered that is \textit{not} adjacent to $X$ (possibly $W = Z$). Take $U_1$ as the first node encountered along $\pi$ with a bidirected edge to $X$ when starting from $W$ in the direction of $X$. Then all other, up to $U_k$ nodes between $U_1$ and $W$ are colliders along $\pi$ with a directed edge into $X$ (again by Lemma 3 and Corollary \ref{corMCIPathCaus}). Similar for some nodes $T$ and $V_1$ for $Y$.
Therefore there exists a path blocked by $Z$ of the form $\pi = X \leftrightarrow U_1 (\leftrightarrow U_2 \leftrightarrow \ldots \leftrightarrow U_k) \aem W \cdots Z \cdots T \mea (V_m \leftrightarrow \ldots \leftrightarrow V_2 \leftrightarrow) V_1 \leftrightarrow Y$ in $\M$, as indicated in Figure \ref{figDSepEdgePath}, with $Z$ as noncollider along the path. Note that $W \in An(\{X,Y\})$ (and similarly for $T$): if $W \tea U_k$ on $\pi$ this is immediate, and if $W \aea U_k$ on $\pi$, then $W$ is not part of an undirected edge, which in combination with Corollary \ref{corMCIPathCaus} leads to $W \in An(\{X,Y\})$.

With generic path $\pi$ we can prove statements (1)-(6), equating $U_1$ with $U$ and $V_1$ with $V$:

(1) By construction, we have $X \leftrightarrow U_1$ and $V_1 \leftrightarrow Y$ along $\pi$. 

(2) By Corollary \ref{corMCIPathCaus}, all nodes along $\pi$ are in $Ant(\{X,Y\})$. As both $U_1$ and $V_1$ are colliders along $\pi$ this reduces to $\{U_1,V_1\} \subset An(\{X,Y\})$. The bidirected edge $X \leftrightarrow U_1$ implies $U_1 \notin An(X)$, and so: $U_1 \in An(Y)$; vice versa for $V_1 \in An(X)$.

(3) If $U_1 \in An(V_1)$, then $V_1 \in An(X)$ and transitivity would imply $U_1 \in An(X)$, contrary the bidirected edge $X \leftrightarrow U_1$, and so $U_1 \notin An(V_1)$; idem for $V_1$ and $Y$.

(4) For the in/dependence relations on $\pi$: given that $X$ and $W$ are not adjacent, they are separated by some minimal set $\bfZ_{XW}$ (not to be confused with $\bfZ_{AA}$ or $\bfZ_{DS}$). By construction, all $\{U_2,\dots,U_k\}$ are part of this set: $U_k$ is needed to block the path $X \leftarrow U_k \aem W$. Conditioning on $U_k$ unblocks the path $X \leftarrow U_{k-1} \leftrightarrow U_k \aem W$ so $U_{k-1}$ is also needed, etc., all the way up to and including $U_2$ (but not $U_1$). As this holds for \textit{any} (minimal) set $\bfZ_{XW}$ that can separate $X$ and $W$, it means there are unblocked paths \textit{into} $U_1$ from both $X$ and $W$ given $\bfZ_{XW}$, and so then conditioning on $U_1$ will make $X$ and $W$ dependent, i.e., \mbox{$\mcdd{X}{W}{\bfZ_{XW}}{U_1}$}. As $Y$ is a descendant of $U_1$, it also implies \mbox{$\mcdd{X}{W}{\bfZ_{XW}}{Y}$}.

(5) Idem \mbox{$\mcdd{Y}{T}{\bfZ_{YT}}{V_1}$} and \mbox{$\mcdd{Y}{T}{\bfZ_{YT}}{X}$}.

(6) Finally, $U_1$ and $V_1$ cannot be adjacent in $\M$: they cannot be connected by a bidirected edge, for that would make the path $\path{X,U_1,V_1,Y}$ unblocked given $\bfZ$; 
by (3) they cannot be connected by an edge $U_1 \rightarrow V_1$ or $U_1 \leftarrow V_1$; and they cannot be connected by an undirected edge because they are both colliders along $\pi$. Therefore $U_1$ and $V_1$ are conditionally independent given some minimal set $\bfZ_{UV}$. For any such minimal separating set $\bfZ_{UV}$, no descendant of $U_1$ or $V_1$ (including $X$ and $Y$) can be part of it, for that would imply either $U_1$ or $V_1$ was ancestor of the other. Including $X$ or $Y$ in the conditioning set would make them dependent given that both $X$ and $Y$ have unblocked paths to $U_1$ and $V_1$ given $\bfZ_{UV}$. 
Therefore, we can find both \mbox{$\mcdd{U_1}{V_1}{\bfZ_{UV}}{X}$} and \mbox{$\mcdd{U_1}{V_1}{\bfZ_{UV}}{Y}$}.
\end{proof}
\end{lem}
By Lemma 2, rule (1), each node in Lemma \ref{lemPD_CausID} that destroys one of the three independences cannot be anterior to any node in that independence, and so leads to identifiable invariant edge-marks (arrowheads).

To make this more precise we first introduce the following definitions:	
\begin{dfn}
A \textbf{minimal independence set} $\I(\M)$ is a set of minimal independencies consistent with a MAG $\M$.
It is called a \textbf{minimal independence model} if it contains at least one separating set for each pair of nonadjacent nodes in the MAG $\M$.
\end{dfn}
The skeleton $\skel$ implied by a minimal independence set $\I(\M)$ corresponds to the undirected graph with no edges between any $(X,Y): \mci{X}{Y}{\bfZ} \in \I(\M)$.
Note that a minimal independence \textit{model} $\I(\M)$ uniquely identifies the Markov equivalence class of $\M$. 

\begin{dfn}  \label{dfnAugmSkel}
Let $\skel$ be the skeleton implied by a minimal independence set $\I(\M)$. Then the  \textbf{Augmented Skeleton} $\skel^+$ is obtained by adding invariant arrowheads at all nodes $W$ on edges to $\{X,Y\} \cup \bfZ$ in $\skel$ that create a single node minimal dependence $\mcdd{X}{Y}{\bfZ}{W}$, for all $\mci{X}{Y}{\bfZ} \in \I(\M)$.
\end{dfn}
Augmentation boils down to repeated application of Lemma 2, rule (1).

From now on we assume that $\I(\skel)$ represents a minimal independence set as output by the PC algorithm with possible addition of one or more \textit{D}-separating sets, consistent with a MAG $\M$. We also assume that we can query an independence oracle for the subsequent dependencies. This implies that the corresponding skeleton $\skel$ matches the skeleton of $\M$, except that it may contain zero, one, or more additional (undirected) edges that all correspond to \textit{D}-sep links in $\M$.
For \textit{D}-sep links in the corresponding augmented skeleton $\skel^+$ this leads to the following pattern:

\textbf{Lemma 4.}
Let $\skel^+$ be the augmented skeleton obtained from a minimal independence set $\I(\skel)$ consistent with a MAG $\M$, such that the only additional edges in $\skel^+$ that do not correspond with an edge in $\M$ are \textit{D}-sep links.
Let $(X,Y)$ be an edge in $\skel^+$ corresponding to a \textit{D}-sep link in the MAG $\M$. 
If there are no (additional) edges in $\skel^+$ between other \textit{D}-separable pairs of nodes in $An(\{X,Y\})$, then $\skel^+$ contains the following pattern: 
\begin{enumerate}
\item[(1)] $U \leftrightarrow X \leftrightarrow Y \leftrightarrow V$ in $\skel^+$, 
\item[(2)] $U$ and $V$ not adjacent in $\skel^+$, 
\item[(3)] paths $V \cdots \rightarrow X$ and $U \cdots \rightarrow Y$ that do not contain arrowheads in the direction of $V$, resp.\ $U$.
\end{enumerate}
\begin{proof}
Follows from Lemma \ref{lemPD_CausID}. 

(1) As $X$ and $U$ are adjacent in $\M$, they are also adjacent in $\skel^+$. Similarly for $Y$ and $V$. Nodes $X$ and $Y$ are also (still) presumed to be adjacent in $\skel^+$. The assumption `no edges in $\skel^+$ between other \textit{D}-sep links in $An(\{X,Y\})$' ensures that the three non-adjacencies (4)-(6) in Lemma \ref{lemPD_CausID} are present in $\I(\M)$; the six subsequent dependences in Lemma \ref{lemPD_CausID} are found by the augmentation procedure, each time adding arrowheads to the corresponding edge. Ultimately this means that $\skel^+$ contains the invariant pattern: $U \leftrightarrow X \leftrightarrow Y \leftrightarrow V$.

(2) In particular (6) in Lemma \ref{lemPD_CausID} ensures that $U$ and $V$ are not adjacent in $\M$. The assumption `no edges in $\skel^+$ between other \textit{D}-sep links in $An(\{X,Y\})$' ensures that $(U,V)$ are not adjacent in $\skel^+$ either.

(3) As $V \in An(X)$ there has to be a path from $V$ in $\skel^+$ that can be(come) oriented as a directed path into $X$. This means the augmentation procedure cannot add an invariant arrowhead in the opposite direction; idem for $U \in An(Y)$.
\end{proof}

The following result generalizes Lemma 5 in the original article as a minimal separating set $\mci{X}{Y}{\bfZ}$ automatically implies $\bfZ \subset Ant(\{X,Y\})_{\setminus \{X,Y\}}$.

\textbf{Lemma 5.}
Let $(X,Y)$ and $(U,V)$ be two possibly overlapping but nonidentical pairs of \textit{D}-separable nodes in a MAG $\M$. If $\{X,Y\} \subset An(\{U,V\})$, then $\{U,V\} \nsubseteq An(\{X,Y\})$.
\begin{proof}
Suppose $U \in An(X)$. If $U \neq X$ then by the given and acylicity $X \in An(V)$, which by transitivity implies $U \in An(V)$, contrary Lemma 3 rule (1). Idem for $U \in An(Y)$. So either $U \in \{X,Y\}$ or $U \notin An(\{X,Y\})$. Idem for $V$. But if both $U \in \{X,Y\}$ and $V \in \{X,Y\}$, with $U \neq V$ in a \textit{D}-sep link, then the two \textit{D}-separable pairs would be identical. Therefore at least one is not ancestor of $\{X,Y\}$, and so $\{U,V\} \nsubseteq An(\{X,Y\})$.
\end{proof}
So two \textit{D}-separable node pairs cannot both be present in each others \textit{D}-separating set. In fact, the ancestor relation induces a partial order over the \textit{D}-sep links:

\begin{lem} \label{lemDSepPOrd} 
Let $\Phi = \big\{\{U_1,V_1\},.., \{U_n,V_n\}\big\}$ be a set of distinct (but not necessarily disjoint) \textit{D}-sep links in a MAG $\M$. Then the relation $\{U_i,V_i\} \preceq \{U_j,V_j\} \iff \{U_i,V_i\} \subseteq An(\{U_j,V_j\})$ defines a partial order over $\Phi$.
\begin{proof}
For all $\{U_i,V_i\},\{U_j,V_j\},\{U_k,V_k\} \in \Phi$:
\begin{enumerate}
  \item Reflexivity: ($\{U_i,V_i\} \preceq \{U_i,V_i\}$) is trivial.
  \item Antisymmetry: (if $\{U_i,V_i\} \preceq \{U_j,V_j\}$ and $\{U_j,V_j\} \preceq \{U_i,V_i\}$ then $\{U_i,V_i\} = \{U_j, V_j\}$) follows from Lemma 5.
  \item Transitivity: (if $\{U_i,V_i\} \preceq \{U_j,V_j\}$ and $\{U_j,V_j\} \preceq \{U_k,V_k\}$, then $\{U_i,V_i\} \preceq \{U_k,V_k\}$) follows from transitivity of the ancestor relationship of nodes in a MAG.
\end{enumerate}
This implies the relation $\preceq$ satisfies the conditions of a partial order over the elements in $\Phi$.
\end{proof}
\end{lem}

As a result, in every non-empty (sub)set of \textit{D}-separable node pairs there is at least one pair that does not have both nodes of any of the other pairs in its ancestors:
\begin{lem} \label{lemDSepOrdMin} 
If $\Phi = \big\{\{U_1,V_1\},.., \{U_n,V_n\}\big\}$ is a non-empty set of distinct (but not necessarily disjoint) \textit{D}-sep links in a MAG $\M$, then there is a $\{U_i,V_i\} \in \Phi$ such that $\forall j \neq i: \{U_j,V_j\} \nsubseteq Ant(\{U_i,V_i\})$.
\begin{proof}
By Lemma 3 rule (4), \textit{D}-sep nodes are not part of an undirected edge, so the statement reduces to $\forall j \neq i:\{U_j,V_j\} \nsubseteq An(\{U_i,V_i\})$. In terms of the partial order defined in Lemma \ref{lemDSepPOrd} this is equivalent to stating that there exists a minimal element with respect to $\preceq$, i.e., an element $\{U_j,V_j\} \in \Phi$ such that there is no other element $\{U_i,V_i\} \in \Phi$ (with $i \ne j$) that precedes it, i.e., such that $\{U_i, V_i\} \preceq \{U_j, V_j\}$. As any finite partially ordered set has at least one minimal element, this proves the lemma.
\end{proof}
\end{lem}
This means that if there are still one or more unidentified \textit{D}-sep links in the augmented skeleton $\skel^+$, then at least one of these has no unidentified \textit{D}-sep links between any two of its ancestors, and so for that \textit{D}-sep link the bidirected edge pattern of Lemma 4 is guaranteed to appear in $\skel^+$. Therefore we can employ the following search strategy to check for \textit{D}-sep links.

\begin{lem} \label{lemDsepHie} 
In a MAG $\M$, all \textit{D}-sep links can be found by repeatedly (and exclusively) checking an augmented skeleton $\skel^+$ for edges that appear as the middle link of the bidirected triple from Lemma 4, while updating $\skel^+$ for each \textit{D}-sep link found.
\begin{proof}
Let $\skel$ be the skeleton of $\M$, possibly with additional edges in $\skel$ that all correspond to \textit{D}-sep links in $\M$, (e.g.\ as obtained from the PC-search stage in the FCI algorithm).
Let $\skel^+$ be the augmented skeleton of $\skel$ w.r.t.\ minimal (in)dependencies implied by $\M$. Then, as long as there are one or more edges in $\skel^+$ that are not in $\M$, then by Lemma \ref{lemDSepOrdMin} at least one of these edges will have no unidentified \textit{D}-sep links (edges in $\skel$ that are not in $\M$) between its ancestors, and so by Lemma 4 this \textit{D}-sep link will show up in $\skel^+$ as the middle edge of the bidirected triple.
Given a procedure to establish whether or not a candidate edge satisfying the bidirected pattern is a \textit{D}-sep link (e.g., FCI's Possible-D-SEP search), then testing all candidate edges, while updating $\skel^+$ for each \textit{D}-sep link identified (remove edge and compute arrowheads for new bidirected triples) until no more can be found, is guaranteed to find all \textit{D}-sep links. This means that at the end the skeleton of $\skel^+$ matches that of the MAG $\M$, and all arrowheads in $\skel^+$ are also in $\M$.
\end{proof}
\end{lem}
This greatly improves the practical running speed of FCI, as often no or hardly any edges need to be checked (after the augmented skeleton has been constructed), but in itself it is not sufficient to guarantee a reduction of the overall complexity to polynomial time, as even a single edge may still require searching through all subsets of order $N$ nodes. The next section shows how a different search strategy can resolve this problem.

\subsection{Proofs - Capturing the \textit{D}-sep nodes}
In proving some of the Lemmas below we often consider marginal MAGs, i.e.\ MAGs $\M'$ obtained by marginalizing out one or more nodes from a base MAG $\M$ in accordance with the rules in \citep{RichSpir02} (see also section~\ref{sec:terminology} for a definition).

First some properties of unblocked paths in an ancestral graph relative to the adjacent ancestors of \textit{D}-sep link $\{X,Y\}$, used in the proof of Lemma 6.

All paths ultimately blocked by one or more of the \textit{D}-sep nodes are unblocked relative to $\AAxy$. 
\begin{lem} \label{lemAllPathsUnblocked} 
In a MAG $\M$, if $\ci{X}{Y}{\AAxy \cup \bfZ}$ with $\bfZ \subset Ant(\{X,Y\})$, and $\pi$ is a path between $X$ and $Y$ that is unblocked relative to $\AAxy \cup \bfZ'$ for $\bfZ' \subset \bfZ$, then:
\vspace{-0.3cm}
\begin{enumerate}
\item[(1)] all colliders on $\pi$ are in $An(\AAxy)$;
\item[(2)] $\pi$ is unblocked given $\AAxy \cup \bfZ^\ast$, $\forall \bfZ^\ast \subseteq \bfZ'$.
\end{enumerate}
\begin{proof}
(1) A path $\pi$ is unblocked relative to a set $\bfZ$ iff every noncollider along $\pi$ is not in $\bfZ$, and every collider on $\pi$ is ancestor of some node in $\bfZ$. If every noncollider along $\pi$ is not present in $\AAxy \cup \bfZ'$ then they are also not present for a subset $\bfZ^\ast \subseteq \bfZ'$. 
Furthermore, every node $Z \in \bfZ'$ that is a descendant of some collider along $\pi$ is in $An(\{X,Y\})$ (given $\bfZ \subset Ant(\{X,Y\})$ and the arrowhead at $Z$ as descendant of the collider), and so has a directed path to $\{X,Y\}$. This directed path goes via penultimate node $U \in \AAxy$, and so it follows that $Z$, and so by transitivity the colliders along $\pi$ as well, are ancestor of a node in $\AAxy$.

(2) Therefore $\pi$ remains unblocked relative to $\AAxy$ in combination with any subset $\bfZ^\ast \subset \bfZ' \subset Ant(\{X,Y\})$, including $\bfZ^\ast = \varnothing$.
\end{proof}
\end{lem}

Also, paths blocked by \textit{D}-sep nodes correspond to sequences of bidirected edges in marginal MAGs.
\begin{lem} \label{lemDseppathBidirSeq}
  In a MAG $\M$ with \textit{D}-separable $X$ and $Y$, if $\mcii{X}{Y}{\AAxy \cup \bfZ}{Z}$ with $\bfZ \subset Ant(\{X,Y\})$, then a path $\pi$ between $X$ and $Y$ in $\M$ that is unblocked relative to $\AAxy \cup \bfZ$ corresponds to a sequence of three or more bidirected edges connecting $X$ and $Y$ in all marginal MAGs $\M'$ over $\{X,Y\} \cup \AAxy \cup \bfZ'$, with $\bfZ' \subseteq \bfZ$.
\begin{proof}
Below we first construct a sequence of unblocked treks in the MAG $\M$ between nodes in $\{X,Y\} \cup \AAxy \cup \bfZ'$ that connects $X$ and $Y$ (\textit{steps 1-3}).
Then we map this sequence to the bidirected edge path in the marginal MAG $\M'$ (\textit{steps 4-6}).
Let $\pi$ be a path in $\M$ between \textit{D}-separable $(X,Y)$ that is unblocked relative to $\AAxy \cup \bfZ$. 

\textit{Step 1: map $\pi$ to sequence $\sigma_U$ of unblocked treks in $\M$.}\\
Let $U_1, \dots, U_m$ be the colliders in $\M$ along the path $\pi$ blocked by $Z$. By Lemma \ref{lemAllPathsUnblocked}-(1), all colliders $U_i \in An(\AAxy)$.
Using similar reasoning as in the beginning of the proof of Lemma \ref{lemPD_CausID}, the path $\pi$ blocked by $Z$ (which is nonadjacent to $X$ and $Y$) must be of the form $X \leftrightarrow U_1 \aem \dots \mea U_m \leftrightarrow Y$.
Each successive pair of colliders $(U_i,U_{i+1})$ along unblocked $\pi$ must be connected by a trek (possibly a single edge $\leftrightarrow$) that does not contain any node in $\AAxy \cup \bfZ$, and so $\sigma_U = [X,U_1,\dots,U_m,Y]$ corresponds to a sequence of treks connecting $X$ and $Y$ in $\M$ that are unblocked relative to any subset of $\AAxy \cup \bfZ$.  

\textit{Step 2: map $\sigma_U$ to sequence $\sigma_V$ of unblocked treks between nodes in $\{X,Y\} \cup \AAxy \cup \bfZ'$.}\\
By Lemma \ref{lemAllPathsUnblocked} the path $\pi$ in $\M$ is also unblocked given $\AAxy \cup \bfZ'$, for any subset $\bfZ' \subseteq \bfZ$, and each collider $U_i$ along $\pi$ is in $An(\AAxy)$. 
For each $U_i$ let $V_i$ be the first descendant of $U_i$ in $\M$ that is in $\AAxy \cup \bfZ'$ (possibly $U_i = V_i$; in particular, $U_1 = V_1$ and $U_m = V_m$). If there are two or more such descendants (along different paths) then simply pick one of these at random. As a result, in $\M$ there are treks between $V_i$ and $V_{i+1}$, and each such trek is again unblocked given any subset of $\AAxy \cup \bfZ'$. Note that the concatenation of the three treks $V_i \aet \dots \aet U_i$, $U_i \aem \dots \mea U_{i+1}$, $U_{i+1} \tea \dots \tea V_{i+1}$ is not necessarily a trek, as a node may occur more than once. This can be remedied by taking a ``shortcut'' via that node. Note that this node cannot become a collider, as at least one of the occurrences of that node must be on one of the directed paths $V_i \aet \dots \aet U_i$ or $U_{i+1} \tea \dots \tea V_{i+1}$ (because $U_i \aem \dots \mea U_{i+1}$ is a trek), and that means that at least one of the edges at that node will have a tail. The result is a trek in $\M$ between $V_i$ and $V_{i+1}$ that is unblocked given any subset of $\AAxy \cup \bfZ'$.
Therefore $\sigma_V = [X,V_1,\dots,V_m,Y]$ corresponds to a sequence of unblocked treks in $\M$ between nodes in $\{X,Y\} \cup \AAxy \cup \bfZ'$.

\textit{Step 3: map $\sigma_V$ to sequence $\sigma_Z$ of unblocked treks between distinct nodes in $\{X,Y\} \cup \AAxy \cup \bfZ'$ in $\M$.}\\
It is possible that there are duplicates in $\sigma_V$, (i.e.\ $V_i = V_j$ with $i \ne j$), e.g.\ in case a descendant in $\AAxy \cup \bfZ'$ is shared by multiple $U_i$. In that case we can remove all nodes $[V_{i+1},\dots,V_j]$ from the sequence $\sigma_V$ while still keeping a contiguous sequence of unblocked treks between $X$ and $Y$. 
Assume we repeatedly merge such doublets (removing all intermediate nodes) until we are left with a sequence $\sigma_Z = [X,Z_1,\dots,Z_k,Y]$ of distinct nodes $Z_i \in \AAxy \cup \bfZ'$, with $Z_i \neq Z_j$ for $i \ne j$, and where each $(Z_i,Z_{i+1})$ is connected by a trek (with arrows into $Z_i$ and $Z_{i+1}$) in $\M$ that is unblocked given any subset $\{X,Y\} \cup \AAxy \cup \bfZ'$.\\
Note that the mapping $\sigma_V \rightarrow \sigma_Z$ is not necessarily unique, e.g.\ if $\sigma_V = [1,3,4,3,5,4,6,2,6,7]$, then either $\sigma_Z = [1,3,5,4,6,7]$ or $\sigma_Z = [1,3,4,6,7]$ will do.

\textit{Step 4: match sequence $\sigma_Z$ to path $\pi'$ in $\M'$.}\\
As each pair $(Z_i,Z_{i+1})$ in the sequence $\sigma_Z$ is connected by a trek in $\M$ that does not contain any noncolliders that are in $\{X,Y\} \cup \AAxy \cup \bfZ'$, it follows that there is an unblocked path between each such pair given any subset of $\{X,Y\} \cup \AAxy \cup \bfZ'$, and so each pair $(Z_i,Z_{i+1})$ must be adjacent in the marginal MAG $\M'$ over $\{X,Y\} \cup \AAxy \cup \bfZ'$. That means the sequence of nodes $\sigma_Z$ corresponds to a path $\pi' = \path{X,Z_1,\dots,Z_k,Y}$ in $\M'$.
Note that this specific path $\pi'$ is \textit{not} necessarily the unblocked path in $\M'$ relative to $\AAxy \cup \bfZ'$ we are looking for (we construct this next). For example, in Figure \ref{figInducingGroupBidirPath}, the sequence $\sigma_Z = [X,Z_1,Z_i,Z_k,Z_j,Y]$ does indeed correspond to a path $X \leftrightarrow Z_1 \leftrightarrow Z_i \rightarrow Z_k \leftrightarrow Z_j \leftrightarrow Y$ in $\M'$, but this path is not unblocked relative to all other nodes in $\M'$ except $\{X,Y\}$, as $Z_i$ is a noncollider along $\pi'$. However, this induces a bidirected edge $X \leftrightarrow Z_k$ in $\M'$, so that the path $\pi^* = X \leftrightarrow Z_k \leftrightarrow Z_j \leftrightarrow Y$ over a subset of nodes along $\pi'$ does correspond to an unblocked path in $\M'$ relative to all other nodes.

\textit{Step 5: find bidirected edges that span nodes along $\pi'$.}\\
Even though the trek in $\M$ between each $(Z_i,Z_{i+1})$ is \textit{into} both nodes, they are not necessarily connected by a bidirected edge along $\pi'$ in $\M'$, as one may be ancestor of the other in $\M$ (and so also in $\M'$). 
Here we show that this induces bidirected edges in the marginal MAG $\M'$ between other nodes along $\pi'$, such that a bidirected edge path $\pi^*$ over a subset of nodes along $\pi'$ connecting $X$ and $Y$ in $\M'$ remains.
(Note that nodes on $\pi'$ are never connected by an undirected edge, as they all have arrowhead marks in $\M$, being (descendants of) colliders along the original $\pi$.)

\begin{figure}[h]
\begin{center}
\includegraphics[scale=0.5]{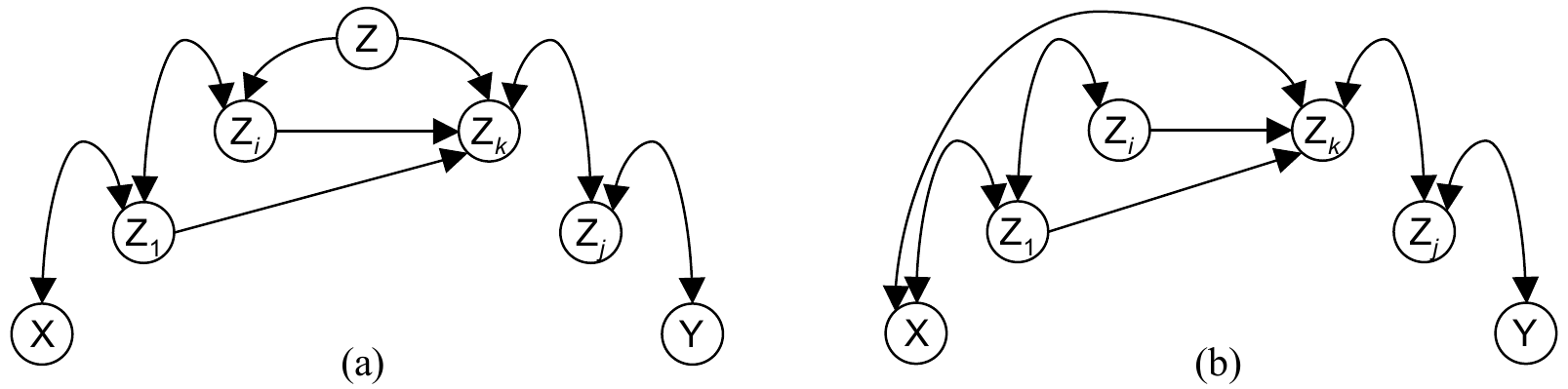}
\caption{\small{(a) Path in $\M$ blocked by \textit{D}-sep node $Z$, (b) Idem in marginal MAG $\M'$ (without $Z$) with corresponding inducing groups $[X,Z_1,Z_i,Z_k],[Z_k,Z_j],[Z_j,Y]$ and induced edge $X \aea Z_k$.}} \label{figInducingGroupBidirPath}
\end{center}
\end{figure}

We now identify \textit{(maximal) inducing groups} of successive nodes $[Z_i,..,Z_j]$ along $\pi'$ that are all ancestor (in $\M$) of the first or the last node (`sink') in the same group, and where two successive inducing groups along $\pi'$ overlap on the last, resp.\ first sink node, so that $[X,..,Z_i],[Z_i,..,Z_j],..,[Z_k,..,Z_m],[Z_m,..,Y]$ along $\pi$.
Each group $[Z_p,..,Z_r]$ is constructed as follows: starting from the last sink node of the previous group $Z_p$ (or $X$ for the first group), add successive nodes along $\pi'$ to the group  until the first node $Z_q$ is encountered that is \textit{not} in $An(Z_p)$. Then, starting from $Y$ back along $\pi'$, find the first node $Z_r$ such that all nodes from $Z_q$ up to $Z_r$ along $\pi'$ are in $An(Z_r)$ in $\M$ (possibly $Z_r = Z_q$). Then $[Z_p,..,Z_r]$ is the next maximal inducing group. 

The two sink nodes of a maximal inducing group are connected by an \textit{inducing path} in $\M$ (hence the name, see also \S1) with respect to nodes not in $\{X,Y\} \cup \AAxy \cup \bfZ'$:
\begin{enumerate}
\item by construction all nodes $\{Z_p,..,Z_r\}$ in the group are in $An(\{Z_p,Z_r\})$,
\item $Z_p$ and $Z_r$ are connected by a path in $\M$ on which all colliders are in $An(\{Z_p,Z_r\})$.
\end{enumerate}
The inducing path between $Z_p$ and $Z_r$ can be constructed as follows. First, one concatenates all the treks $Z_p \aem \dots \mea Z_{p+1}, Z_{p+1} \aem \dots \mea Z_{p+2}, \dots, Z_{r-1} \aem \dots \mea Z_r$. In case a node $Q$ occurs more than once, one takes the shortcut by deleting all intermediate nodes between the left-most and right-most occurrence of $Q$. Such $Q$ can occur at most once on the original path $\pi$, and the other occurrence(s) must be on a directed path to one of the nodes $Z_p,\dots,Z_r$. Therefore, if the remaining node $Q$ becomes a collider when making the shortcut, then $Q \in An(\{Z_p,\dots,Z_r\}) = An(\{Z_p,Z_r\})$.

As a result, by Theorem 4.2-(i) in \cite{RichSpir02} the two sink nodes of a maximal inducing group are connected by an edge in $\M'$. Furthermore, this edge must be a bidirected edge, as sink nodes in a maximal inducing group cannot be ancestor of each other:
\begin{enumerate}
\item $Z_r \notin An(Z_p)$, otherwise by transitivity also $Z_q \in An(Z_p)$, contrary the given, 
\item $Z_p \notin An(Z_r)$, because for $Z_p = X$ (first group) as part of \textit{D}-sep link $(X,Y)$, the given $\AAxy \cup \bfZ \subset Ant(\{X,Y\})$ and acyclicity implies $X \notin An(Z_r) \subset An(\bfZ)$; and if $Z_p$ is the sink node of the previous group $[Z_m,..,Z_p]$, then all nodes $[Z_p,..,Z_r]$ would also satisfy the conditions for inclusion in that group, and we would have obtained $[Z_m,..,Z_r]$, contrary the given. 
\end{enumerate}

\textit{Step 6: obtain bidirected edge path $\pi^\ast$ in $\M'$}.\\
Therefore in $\M'$ there is a path $\pi^\ast$ of bidirected edges connecting $X$ to $Y$ via the sink nodes of neighbouring inducing groups along $\pi'$. The connection between the path $\pi$ in $\M$ and the paths $\pi'$ and $\pi^*$ in $\M'$ is illustrated in Figure \ref{figOverlapMaxIndGroupBidirPath}.

For every $\pi'$ there are at least three distinct inducing groups: the first and last edge along $\pi'$ in $\M'$ (corresponding to $X \leftrightarrow U_1$ resp.\ $U_m \leftrightarrow Y$ along $\pi$ in $\M$, as both $U_1$ and $U_m$ as subset of $\AAxy$ are also in $\M'$) are part of disjoint groups, as by Lemma \ref{lemPD_CausID}-(3) $U_1 \notin An(U_m)$, and vice versa. That means there must be at least one other group to bridge the gap between $U_1$ and $U_m$ from which follows that there are at least \textit{three} such bidirected edges on the path $\pi^\ast$ connecting $X$ and $Y$ in $\M'$. This proves the Lemma.
\end{proof}
\end{lem}

\begin{figure}[h]
\begin{center}
\includegraphics[scale=0.5]{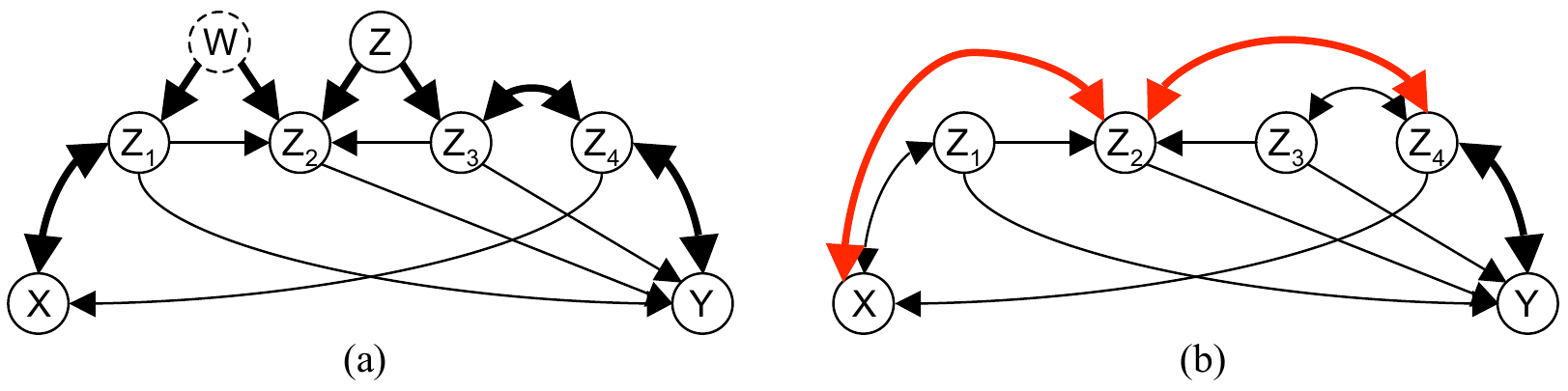}
\caption{\small{Induced bidirected edge path.\\
(a) Path $\pi = \path{X,Z_1,W,Z_2,Z,Z_3,Z_4,Y}$ in $\M$ blocked by \textit{D}-sep node $Z$, (b) Corresponding path $\pi' = \path{X,Z_1,Z_2,Z_3,Z_4,Y}$ in marginal MAG $\M'$ over $\{X,Y\} \cup \AAxy (= \{Z_1,Z_2,Z_3,Z_4\})$, with maximal inducing groups $[X,Z_1,Z_2],[Z_2,Z_3,Z_4],[Z_4,Y]$, leading to induced edges $X \leftrightarrow Z_2$ and $Z_2 \leftrightarrow Z_4$ (red), to give the bidirected edge path $\pi^* = \path{X,Z_2,Z_4,Y}$ in $\M'$ (bold).}} \label{figOverlapMaxIndGroupBidirPath}
\end{center}
\end{figure}

With this we can prove Lemma 6 and Lemma \ref{lemDsepIter} on being able to find all required \textit{D}-sep nodes sequentially as part of a separating set between nodes already found.

All \textit{D}-sep nodes for a pair $(X,Y)$ also appear in another minimal conditional independence: 

\textbf{Lemma 6.} 
In a MAG $\M$, if $Z \in \bfZ$ is a \textit{D}-sep node in $\mci{X}{Y}{\bfZ}$, then $Z$ is also part of a minimal separating set between another pair of nodes from $\{X,Y\} \cup \bfZz \cup \AAxy$, neither of which are part of an undirected edge in $\M$.
\begin{proof}
By Lemma \ref{lemDseppathBidirSeq}, in the marginal MAG $\M'$ over $\{X,Y\} \cup \bfZz \cup \AAxy$ there is a sequence of at least 3 bidirected edges connecting $X$ and $Y$. If this sequence of edges still exists in the MAG $\M^\ast$ over $\{X,Y\} \cup \bfZ \cup \AAxy$, then $X$ and $Y$ are not separated given $\bfZ \cup \AAxy$, contrary the given. Note that any adjacency in $\M^\ast$ that is also present in $\M'$ must have identical arrow/tail-marks in $\M^\ast$ and $\M'$, as both edges must express the same anterior relations in $\M$. Therefore at least one of these edges is eliminated in $\M^\ast$, which implies the existence of a set that can separate these two nodes from $\{X,Y\} \cup \bfZz \cup \AAxy$. \textit{D}-sep node $Z$ is necessarily part of that set, otherwise the edge would already be eliminated in $\M'$.
Given that both separated nodes have arrowhead marks in $\M'$ (as part of a bidirected edge path) it also follows that neither can be part of undirected edge as well.
\end{proof}

\begin{lem} \label{lemDsepIter}
In a MAG $\M$ with \textit{D}-sep link $(X,Y)$, all \textit{D}-sep nodes $\bfZ_{DS}$ in $\mcii{X}{Y}{\AAxy}{\bfZ_{DS}}$ can be found sequentially as part of a minimal separating set between a pair of nodes already found, starting from $\{X,Y\} \cup \AAxy$.
\begin{proof}
Starting from the marginal MAG $\M_0$ over $\{X,Y\} \cup \AAxy$ we infer from Lemma \ref{lemDseppathBidirSeq} that there is a sequence of bidirected edges in $\M_0$ connecting $X$ and $Y$. Analogous to the rationale in Lemma 6: if this sequence of edges still exists in $\M^*$ over $\{X,Y\} \cup \AAxy \cup \bfZ_{DS}$, then $X$ and $Y$ are not separated, contradicting the assumptions. Therefore at least one of these edges must be eliminated in $\M^\ast$, which implies the existence of a minimal separating set containing one or more of the nodes from $\bfZ_{DS}$. If only a subset $\bfZ_1 \subset \bfZ$ of nodes are needed in this separating set, then we can apply the same argument again to the marginal MAG $\M_1$ over $\{X,Y\} \cup \AAxy \cup \bfZ_1$. Each time we find new nodes from $\bfZ$ that are part of a separating set between some pair of nodes we already found, until all have been added.
\end{proof}
\end{lem}

\subsection{Proofs - Building the hierarchy}

Lemma \ref{lemDsepIter} describes a procedure to find all required \textit{D}-sep nodes for a given \textit{D}-sep link $(X,Y)$, starting from the adjacent ancestors of $X$ and $Y$. However, there is a possible snag in the sense that the lemma ensures that each \textit{D}-sep node can be found as part of \textit{some} minimal separating set between a pair of nodes already found, but standard search strategies like PC only look for a \textit{single} minimal sepset between each separable pair of nodes. As a result, if there are multiple possible minimal sepsets it is not a priori guaranteed that the \textit{D}-sep nodes we are looking for are indeed contained in the sepset returned by the PC-search stage.
However, it turns out that they can still be found as part of an `ancestral superset' that is built recursively from results already obtained.

For that we introduce the following recursive definition for a set of separating nodes:
\begin{dfn}
Let $\I$ be a minimal independence set, then for a set $\bfX$ the \textbf{hierarchy} $HIE(\bfX,\I)$ is the union of $\bfX$ and all nodes that appear in a minimal separating set in $\I$ between any pair of nodes in $HIE(\bfX,\I)$.
\end{dfn}
The recursion as formula: let $\bfQ_0 = \bfX$, and $\bfQ_{i+1} = \bfQ_i \cup \left( \bigcup_j \bfW_j: U_j,V_j \in \bfQ_i, \mci{U_j}{V_j}{\bfW_j} \in \I \right )$. Then, if $n$ is the lowest index for which $\bfQ_{n+1} = \bfQ_n$, $HIE(\bfX,\I) = \bfQ_n$, with $n < $ nr.\ of variables in $\I$.
Note that $\bfX \subseteq HIE(\bfX,\I) \subseteq Ant(\bfX)$ for any $\bfX$ and any minimal independence set $\I(\M)$.

The crucial result is now that for an arbitrary minimal independence set the hierarchy of a \textit{D}-sep link and its adjacent ancestors is guaranteed to be a separating superset that contains all required \textit{D}-sep nodes.

\textbf{Lemma 7.}
In a MAG $\M$ with minimal independence model $\I(\M)$, suppose that $X$ and $Y$ are non-adjacent. Let $\I(\skel^+) \subset \I(\M)$ be a subset that contains the same separating sets between ancestors of $X$ and/or $Y$ in $\M$, except that it does not contain a separating set for $\{X,Y\}$.
If $\bfQ = HIE(\{X,Y\} \cup \AAxy,\I(\skel^+))_{\setminus \{X,Y\}}$, then $\bfQ$ is a separating set between $X$ and $Y$, i.e.\ $\ci{X}{Y}{\bfQ}$.
\begin{proof}
As $X$ and $Y$ are non-adjacent in $\M$, there is a minimal separating set $\mci{X}{Y}{\bfZ}$.
By Lemma \ref{lemDsepNodes} and Corollary \ref{corAAsep}, \mbox{$\mci{X}{Y}{\bfZ}$} implies that \mbox{$\mcii{X}{Y}{\bfQ}{\bfZ^*}$}, with $\bfQ \supset \AAxy$ as defined above and where all $\bfZ^* \subset \bfZ$ are \textit{D}-sep nodes for $(X,Y)$. 

If $(X,Y)$ is not a \textit{D}-sep link then $\bfZ^* = \varnothing$, and the previous immediately reduces to $\ci{X}{Y}{\bfQ}$. 

Suppose $\bfZ^* \neq \varnothing$, i.e.\ $\bfQ$ is not a \textit{D}-separating set for $(X,Y)$. Then we can write $\mcii{X}{Y}{\bfQ \cup \bfZ^*_{\setminus Z}}{Z}$, for some $Z \in \bfZ^*$, and so there is an unblocked path $\pi$ in $\M$ relative to $\bfQ \cup \bfZ^*_{\setminus Z}$.

By Lemma \ref{lemDseppathBidirSeq} any path $\pi$ unblocked without $Z$ corresponds to a sequence of three or more bidirected edges in any marginal MAG over $\{X,Y\} \cup \AAxy \cup \bfQ'$ with $\bfQ' \subseteq (\bfQ \cup \bfZ^*_{\setminus Z})$, including the marginal MAG $\M'$ over $\{X,Y\} \cup \bfQ$.

Therefore there is an unblocked path $\pi'$ between $X$ and $Y$ relative to $\bfQ$ in this $\M'$ that consists of a sequence of three or more bidirected edges. If these edges are still present in $\M$, then the path is still unblocked relative to any set that includes $\bfQ$, counter to $\mcii{X}{Y}{\bfQ}{\bfZ^*}$. 
If one of the edges from $\pi'$ is no longer present in $\M$, then that implies the existence of a minimal separating set in $\I(\M)$ between two nodes from $\{X,Y\} \cup \bfQ$. Any such separating set would satisfy the conditions for inclusion in the hierarchy $\bfQ$ (definition 6),
and so already be part of $\bfQ$. But then the corresponding edge should be absent from the marginal MAG $\M'$ over $\{X,Y\} \cup \bfQ$, contrary the assumption that it was part of the path $\pi'$. 

Therefore the assumption that $\bfZ^* \neq \varnothing$ leads to a contradiction, and so $\bfQ$ must already contain all required nodes to form a separating set for $(X,Y)$, i.e.\ $\ci{X}{Y}{\bfQ}$.
\end{proof}

\begin{figure}[h]
\begin{center}
\includegraphics[scale=0.9]{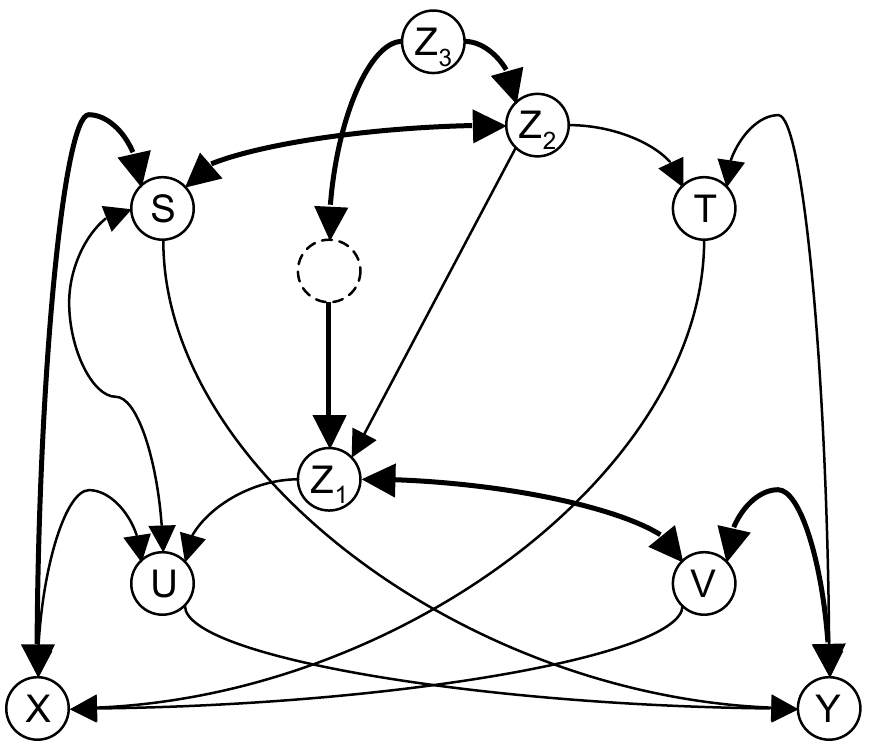}
\caption{\small{Illustration that it is not sufficient to look at separating sets between nodes adjacent to $\{X,Y\}$, but that it is necessary to include the full recursive hierarchy: nodes $X$ and $Y$ are \textit{D}-separated by $\{S,T,U,V,Z_1,Z_2,Z_3\}$, but $Z_3$ (blocking the path in bold) is \textit{only} present (and necessary) in $\mci{S}{Z_1}{Z_2,Z_3}$, with $Z_1 \notin Adj(\{X,Y\})$.}} \label{figHieNecess2}
\end{center}
\end{figure}

The resulting \textit{D}-sep set can be converted into a minimal \textit{D}-separating set in at most $N^2$ additional independence tests, by removing redundant nodes one-by-one until no more can be found, see \citep{TianPP98}. 

While searching for \textit{D}-sep links in the augmented skeleton $\skel^+$ we may not yet know the true adjacent ancestors of a \textit{D}-sep candidate pair $\{X,Y\}$ in the underlying MAG $\M$. 
For a node with degree bounded by $k$, it has to be a combination of at most $k$ from $N$ nodes, which for a pair implies one from worst case order $N^k \times N^k = N^{2k}$ different sets. If there is a \textit{D}-separating set for candidate $\{X,Y\}$, then it is guaranteed to appear in the hierarchy implied by one of these sets. Therefore, in a sparse graph, we need at most a polynomial number of tests (in $N$) to find a \textit{D}-sep set. 

We can now prove the main claim of the article:
\begin{thm}
Let $\M$ be a MAG over $N$ observed nodes corresponding to a distribution that is faithful to some underlying causal DAG $\G$, such that the node degree in $\M$ is bounded by some constant $k$, then the sound and complete equivalence class PAG $\cP$ can be obtained from worst case polynomial order $N^{2(k+2)}$ independence tests, even when latent variables and selection bias may be present. 
\begin{proof}
Follows from a combination of (known) complexity results for the three main stages required to obtain the PAG $\cP$:
\vspace{-0.2cm}
\begin{itemize*}
\item[1.] find PC-skeleton graph $\skel$ from adjacency search,
\item[2.] eliminate \textit{D}-sep links to obtain the skeleton of $\M$,
\item[3.] orient invariant edge marks to obtain PAG $\cP$.
\end{itemize*}
\vspace{-0.2cm}
The first stage is known to require worst case order $N^{k+2}$ independence tests, as it searches for subsets $\leq k$ from $N-2$ variables in order to eliminate at most $N(N-1)$ edges \citep{SGS00}.

This article has shown that the second stage can be completed in a number of independence tests that is also worst case polynomial order in $N$. First it suffices to find/update the augmented skeleton $\skel^+$ (Definition 4) in order to obtain up to $N^2$ possible candidate \textit{D}-sep links (Lemma 4). Augmenting the PC-graph may take order $N^3$ tests as it needs to check for at most $N-2$ nodes for $N(N-1)$ edges eliminated so far.
Lemma 16 ensures that as long as not all \textit{D}-sep links have been found then at least one of these has no unidentified \textit{D}-sep links between its ancestors and so can be identified as a bidirected edge triple in $\skel^+$ (Lemma 4).
For each possible D-sep link $\{X,Y\}$ we need to find the set of possible adjacent ancestors $\AAxy$ (Definition 3) in $\M$ to compute the corresponding hierarchy (Lemma 7). For both candidates $\{X,Y\}$ this implies searching for at most $k$ nodes from $N-2$ possible neighbours, which boils down to worst case $N^k \times N^k = N^{2k}$ independence tests. On finding a \textit{D}-sep set we may need order $N^2$ tests to convert it into a minimal separating set \citep{TianPP98}, update the augmented skeleton ($N^3$), and possibly recheck up to $N^2$ previously tried-but-failed \textit{D}-sep links. This leads to an overall complexity of the second stage of worst case order $N^2 \times N^{2k} \times N^2 = N^{2(k+2)}$, where augmenting the graph and conversion into minimal \textit{D}-sep sets do not contribute to the leading order terms.

The third stage does not require additional independence tests at all. Complexity of the orientation rules lies mainly in checking for each edge mark for existence of certain paths in $\cP$, which can be done in order $N^2k$ by a generic `reachability' algorithm, with $Nk \sim$ nr.\ of edges. As there are $N^2$ possible edge marks, this overall complexity is worst case order $N^4k$.

As the first and third stage do not contribute to the leading order terms from the second stage, it implies that the overall complexity of finding the sound and complete PAG takes at most order $N^{2(k+2)}$ independence tests. 
\end{proof}
\end{thm}
Note that in practice the typical performance is much better than this worst case result suggests: relatively few \textit{D}-sep candidates with even fewer \textit{D}-sep links. This limits the nr.\ of adjacent nodes per candidate pair in the second stage to $\sim k$, which reduces the most expensive term from $N^{2k}$ down to constant order $2^{2k}$ (twice all subsets from $k$ nodes), independent of $N$.

A description of the FCI+ algorithm that implements this result can be found in the main article.

\subsubsection*{Acknowledgements} 
This research was supported by the NWO (Netherlands Organization for Scientic Research), grant nr.\ 612.001.202.
JM was supported by NWO VENI grant nr.\ 639.031.036.
\bibliographystyle{plainnat}
\bibliography{UAI2013_arXiv_references}
 
\end{document}